\documentclass[10pt,twocolumn,letterpaper]{article}

\usepackage[pagenumbers]{cvpr} 
\usepackage{algorithmicx} %
\usepackage{algpseudocode} %
\usepackage[table]{xcolor} %
\usepackage{graphicx}
\usepackage{microtype}
\usepackage{algorithm}
\usepackage{booktabs}
\usepackage{amsmath}
\usepackage{multirow,tabularx,multicol}
\usepackage{relsize}
\definecolor{aliceblue}{rgb}{0.94, 0.97, 1.0}

\usepackage{microtype}

\renewcommand{\paragraph}[1]{\vspace{.5em}\noindent\textbf{#1.}}

\usepackage{soul}
\setuldepth{foobar}

\definecolor{cvprblue}{rgb}{0.21,0.49,0.74}
\usepackage[pagebackref,breaklinks,colorlinks,allcolors=cvprblue]{hyperref}

\def\paperID{18983} 
\def\confName{CVPR}
\def\confYear{2026}

\title{DiFlowDubber: Discrete Flow Matching for Automated Video Dubbing via Cross-Modal Alignment and Synchronization}

\author{Ngoc-Son Nguyen$^{1}$ \quad Thanh V. T. Tran$^1$ \quad Jeongsoo Choi$^2$ \quad Hieu-Nghia Huynh-Nguyen$^{1*}$\\ Truong-Son Hy$^3$ \quad Van Nguyen$^{1\dagger}$\\
$^1$FPT Software AI Center, Vietnam \quad
$^2$KAIST, South Korea \\
$^3$University of Alabama at Birmingham, USA 
\\
{\small \url{https://nngocson2002.github.io/projects/diflowdubber}}
\vspace{-0.5cm}
}

\begin{document}
\maketitle
\let\thefootnote\relax\footnotetext{$^*$Independent researcher, work done in part while author was at the FPT Software AI Center, Vietnam. \hspace{2pt}$^\dagger$Corresponding author.}
\begin{abstract}
Video dubbing requires content accuracy, expressive prosody, high-quality acoustics, and precise lip synchronization, yet existing approaches struggle on all four fronts. To address these issues, we propose \textbf{DiFlowDubber}, the first video dubbing framework built upon a discrete flow matching backbone with a novel two-stage training strategy. In the first stage, a zero-shot text-to-speech (TTS) system is pre-trained on large-scale corpora, where a deterministic architecture captures linguistic structures, and the Discrete Flow-based Prosody-Acoustic (DFPA) module models expressive prosody and realistic acoustic characteristics. In the second stage, we propose the Content-Consistent Temporal Adaptation (CCTA) to transfer TTS knowledge to the dubbing domain: its Synchronizer enforces cross-modal alignment for lip-synchronized speech. Complementarily, the Face-to-Prosody Mapper (FaPro) conditions prosody on facial expressions, whose outputs are then fused with those of the Synchronizer to construct rich, fine-grained multimodal embeddings that capture prosody-content correlations, guiding the DFPA to generate expressive prosody and acoustic tokens for content-consistent speech. Experiments on two benchmark datasets demonstrate that DiFlowDubber outperforms prior methods across multiple evaluation metrics.
\end{abstract}
\vspace{-9pt}
\section{Introduction} \label{sec:introduction}
\begin{figure}[t]
\vspace{-0.2cm}
\centering
\includegraphics[width=\columnwidth]{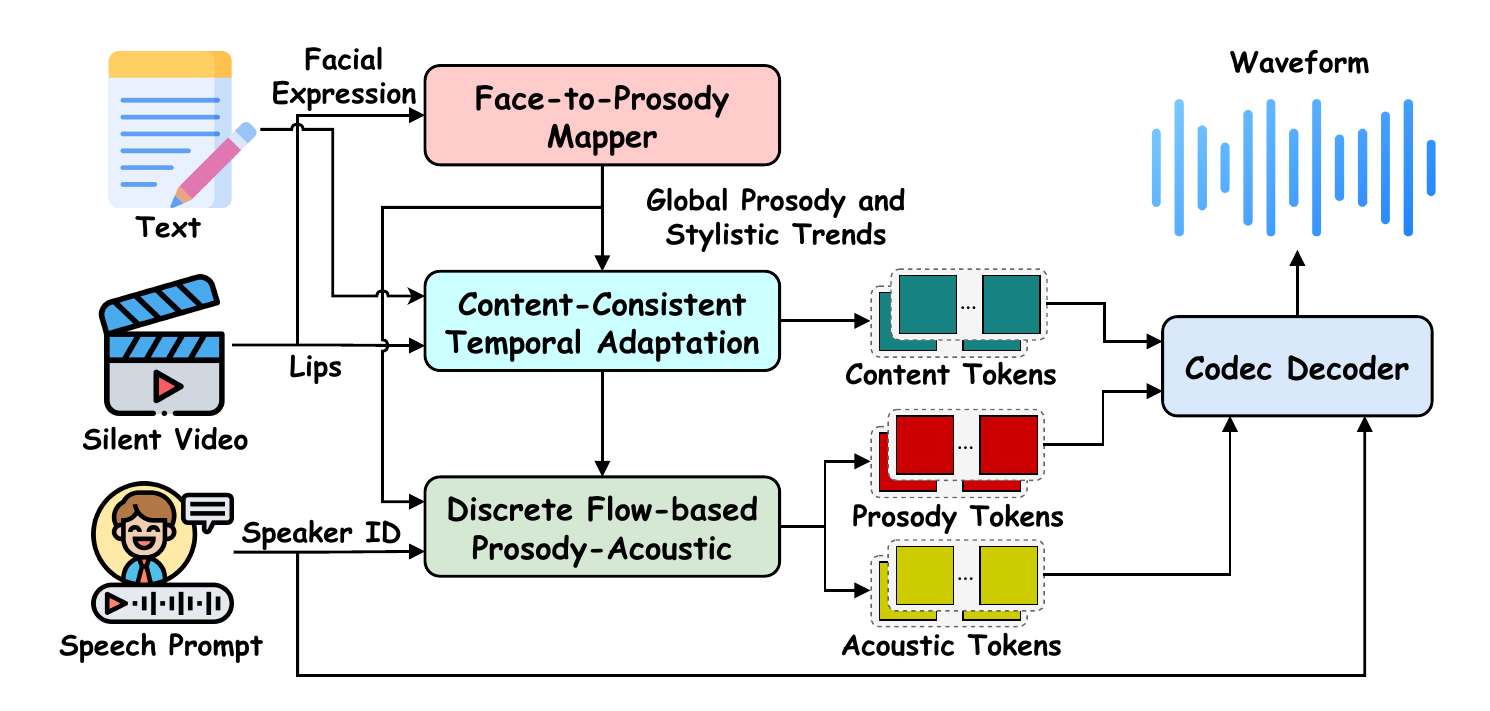}
\caption{\textbf{Overall Inference Pipeline of DiFlowDubber.} \textit{Face-to-Prosody Mapper} module predicts global prosody priors that capture global prosody and stylistic cues from facial expressions. \textit{Content-Consistent Temporal Adaptation} module generates discrete content tokens conditioned on lip movements, text, and prosody priors, ensuring consistency with the target text transcription and temporal alignment. \textit{Discrete Flow-based Prosody-Acoustic} module generates diverse yet globally consistent prosody tokens under the guidance of the prosody prior, together with corresponding acoustic tokens. The speech waveform is synthesized from the predicted tokens and speaker embedding via a Codec Decoder.
}
\label{fig:overview}
\vspace{-15pt}
\end{figure}
Visual voice cloning (V2C), or video dubbing, generates speech conditioned on subtitle text, reference speech, and reference video, ensuring the target voice is preserved, lip-synchronized, and expressive of video-driven emotion.
Despite notable advances in video dubbing~\citep{hpmdubbing, styledubber, speaker2dubber, produbber, emodubber}, synthesized speech often falls short in overall quality, particularly in terms of naturalness, pronunciation accuracy, and synchronization. A primary bottleneck is the scarcity of large-scale legally shareable dubbing data, which is constrained by copyright issues. Meanwhile, modern TTS systems pre-trained in large-scale high-quality corpora \citep{libritts} are capable of generating high-fidelity speech, yet they still lack video-conditioned affect and precise lip synchronization. A natural solution is to adapt pre-trained TTS models for video dubbing while preserving their original capabilities. Speaker2Dubber \citep{speaker2dubber} attempts this approach by pre-training on a large TTS corpus and then adapting its phoneme encoder for the dubbing task. While this improves the richness of linguistic representations, it fails to fully leverage the advantages of large-scale TTS pre-training, such as prosody and acoustic modeling. To address this limitation, ProDubber \citep{produbber} introduces a prosody-enhanced acoustic pre-training stage and an acoustic-prosody disentanglement stage to achieve higher-quality dubbing with better prosody alignment. However, this method primarily focuses on improving speech quality but still struggles with synchronization, as it relies on a duration predictor to estimate lip movements that are not constrained by the actual video length, resulting in poor audio-visual synchronization.

To address these issues, we present \textbf{DiFlowDubber}, a novel approach for automated video dubbing based on Discrete Flow Matching (DFM) \citep{dfm}. Our motivation is to design a framework that fully exploits the advantages of large-scale TTS pre-training, particularly in modeling prosody, content, and acoustic characteristics, and adapt it for video dubbing while preserving pronunciation accuracy, naturalness, and precise lip synchronization. To this end, we first employ a pre-trained FACodec \citep{ns3} to decompose speech into prosody, content, and acoustic tokens, which serve as our target discrete representations. This choice is motivated by two key factors: \textbf{(1)} FACodec factorizes speech attributes such as prosody, content, and acoustic information, making them easier to control and model; and \textbf{(2)} it is pre-trained on a large-scale multi-speaker corpus, providing a strong and reliable codec foundation for our target data. Building on this representation, we design a novel two-stage pipeline:

\textbf{In the first stage}, we develop a zero-shot TTS system that models factorized speech representations to synthesize high-quality speech in a discrete setting. To this end, the system is pre-trained in a zero-shot TTS configuration, where we employ a simple deterministic content modeling architecture to efficiently capture linguistic structure. In contrast, for prosody and acoustic attributes, we adopt a generative modeling approach using discrete flow matching through the \textit{\textbf{D}iscrete \textbf{F}low-based \textbf{P}rosody-\textbf{A}coustic} (DFPA) module to capture expressive prosodic variations and realistic acoustic diversity from the corpus. \textbf{In the second stage}, the pre-trained TTS model is adapted for video dubbing, as illustrated in Figure~\ref{fig:overview}. In particular, we first design the \textit{\textbf{Fa}ce-to-\textbf{Pro}sody Mapper} (FaPro) module to extract global prosodic cues from face-cropped video frames. In addition, we introduce the \textit{\textbf{C}ontent-\textbf{C}onsistent \textbf{T}emporal \textbf{A}daptation} (CCTA) module, which transfers knowledge from the TTS domain to generate content representations consistent with the target text transcription. Central to this module is the Synchronizer, which replaces the duration predictor of the TTS model and takes phoneme sequences and lip-cropped visual features as inputs to capture temporally aligned content details, thereby addressing the synchronization challenge during domain adaptation. The global prosody prior from FaPro and the synchronized content representation from the Synchronizer module are then fused to construct rich, fine-grained multimodal embeddings that capture prosody-content correlations. Finally, this embedding is used to predict content token sequences and, together with the global prosody features, is fed into the DFPA, which is initialized from the pre-trained TTS model and fine-tuned to generate diverse prosody and acoustic tokens. We summarize our contributions as follows:
\begin{itemize}
    \item We propose DiFlowDubber, the first end-to-end automated video dubbing framework built upon a discrete flow matching generative backbone. Our approach consists of a zero-shot TTS pre-training stage that fully exploits content information, expressive prosodic variations, and realistic acoustic diversity from large-scale corpora, followed by a transfer learning stage that adapts the pretrained TTS model to visual inputs for video dubbing.
    \item We propose the CCTA module for effective knowledge transfer from the TTS domain to the video dubbing setting. Central to CCTA is the \textit{Synchronizer}, which replaces the duration predictor of the TTS model with a dual-alignment mechanism over text, video, and speech, enforcing precise cross-modal synchronization while preserving content consistency with the target transcription. Complementarily, the FaPro extracts global prosodic cues from facial expressions to guide expressive speech generation.

    \item Extensive experiments on two major benchmarks demonstrate that DiFlowDubber surpasses state-of-the-art (SOTA) baselines, achieving superior content accuracy, naturalness, and synchronization with video.
\end{itemize}

\vspace{-5pt}
\section{Related work} \label{sec:related-work}
\subsection{Speech synthesis} 
Speech synthesis has undergone rapid progress in recent years, moving from concatenative and statistical parametric approaches \citep{concatenative_tts,hmm_tts} to deep generative models that achieve human-like naturalness. Early neural architectures \citep{tacotron,8461368} demonstrated the feasibility of end-to-end TTS by jointly modeling prosody and waveform generation. Subsequent advances have focused on improving efficiency and controllability, with non-autoregressive models \citep{fastspeech,glow_tts,fastspeech2} reducing inference latency while maintaining quality. More recently, discrete representation learning \citep{wav2vec2,resynthesis_ssl,ns3,valle,wang2025maskgct} and diffusion/flow-based methods \citep{glow_tts, voiceflow,f5tts,ozspeech,diflowtts, kang23_interspeech, ns2, ns3, lee2025dittotts, kim2023pflow, le2023voicebox, matcha-tts, e2tts} have further improved expressiveness and robustness, allowing fine-grained control over prosody, speaker identity, and style. Parallel to these developments, large language models have been integrated with TTS pipelines \citep{llm_tts_survey,fang2025llamaomni, valle-x, valle-r, valle2, ella-v, melle, voicecraft, mobilespeech, spark-tts, valle}, bridging the gap between linguistic understanding and acoustic realization. Despite these advances, most systems remain text-driven, limiting their applicability in multimodal scenarios such as video dubbing, where visual articulatory cues are essential for precise synchronization. Building on these foundations, we extend discrete generative modeling to incorporate visual cues, addressing the modality gap between text-based TTS and video-conditioned speech synthesis.

\subsection{Visual Voice Cloning}

The V2C task requires the models to align the prosody and timing of generated speech with the input video to achieve automatic dubbing. Early efforts in V2C explored lip-to-speech generation \citep{intelligile_l2s,yemini2024lipvoicer,resound}, showing that visual articulatory cues provide strong constraints for intelligibility. More recent approaches have incorporated multimodal inputs, leveraging text and video to improve synchronization and naturalness. One line of work has emphasized hierarchical modeling, where prosodic features such as rhythm, intonation, and duration are jointly synchronized with facial expressions and contextual video dynamics \citep{hpmdubbing}. Other approaches have focused on character-specific expressiveness: for example, StyleDubber \citep{styledubber} integrates phoneme-level details with emotion variation to reproduce distinctive speaking styles, while EmoDubber \citep{emodubber} leverages a flow-based design with emotion-aware conditioning to modulate intensity. Addressing pronunciation accuracy, Speaker2Dubber \citep{speaker2dubber} pre-trains its phoneme encoder on large-scale text-speech corpora to improve articulation. Among the most relevant for our study, ProDubber \citep{produbber} adopts a two-phase training pipeline that transfers knowledge from extensive TTS datasets. However, ProDubber employs a duration predictor to upsample acoustic text features to the spectrogram length based on predicted alignments. Since this process is not constrained by the actual video length, it can lead to suboptimal lip-sync error (LSE) scores (see \Cref{tab:overall-results-grid,tab:overall-results-chem}).
\vspace{-0.1cm}
\section{Methodology} \label{sec:method}
\begin{figure*}[t]
\vspace{-0.4cm}
\centering
\includegraphics[width=0.98\textwidth]{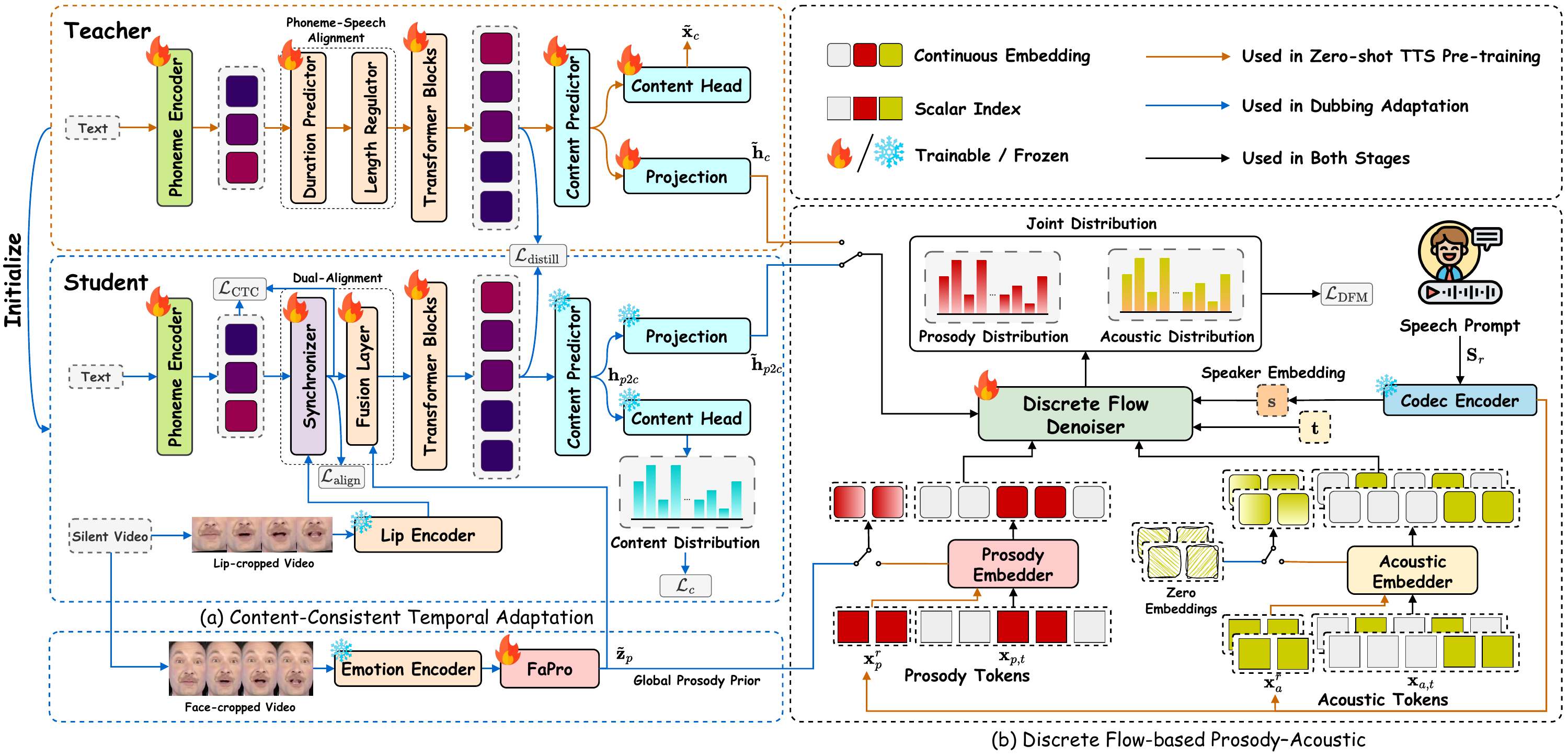} 
\caption{\textbf{Detailed of DiFlowDubber Pipeline.} Our framework is a two-stage pipeline. The first stage performs zero-shot TTS pre-training, where a simple deterministic content modeling architecture efficiently captures linguistic structures (orange dashed box). For prosody and acoustic attributes, we adopt \textbf{(b)} \textit{Discrete Flow-based Prosody-Acoustic} (DFPA) module to model expressive prosodic variations and realistic acoustic diversity from the corpus. In the second stage, the model is adapted to the video dubbing task, where \textbf{(a)} \textit{Content-Consistent Temporal Adaptation} module transfers consistent content knowledge from the TTS domain and generates temporally aligned content representations. Meanwhile, FaPro extracts a global prosody prior from facial expression cues. The DFPA module then models the joint distribution of prosody and acoustic tokens conditioned on the global prosody prior, latent content representations, and speaker embedding.}
\label{fig:framework}
\vspace{-0.4cm}
\end{figure*}

\subsection{Overview}
Our framework is designed as an end-to-end pipeline. We first pre-train a zero-shot TTS system that learns to control and model complex speech attributes, including prosody, content, and acoustic characteristics. The content attribute is modeled as illustrated in Figure \ref{fig:framework} (Teacher box), while the DFPA generates tokens that capture expressive prosody and realistic acoustic characteristics consistent with a given speech prompt (Figure \ref{fig:framework}b, orange arrow). We then adapt this pre-trained model to the video dubbing domain, enabling it to inherit the advantages of the large-scale TTS corpus by integrating visual cues. This adaptation is achieved through two key components:
\textbf{(a)} CCTA (Section \ref{sec:ccta}), which transfers knowledge from the TTS domain to generate content representations consistent with the target text transcription, achieving precise lip synchronization and capturing global prosodic states conveyed through visual cues; and \textbf{(b)} DFPA modeling with modified conditioning to better suit the visual modality which is detailed in Section \ref{sec:DFPA_adapt}.
\vspace{-2pt}
\subsection{Data Preparation}
\vspace{-2pt}
\paragraph{Discrete Speech Tokens} We leverage the pre-trained FACodec \cite{ns3}, which consists of a Codec Encoder and a Codec Decoder. Given a speech waveform $\mathbf{S}$, we apply the Codec Encoder to decompose $\mathbf{S}$ into discrete prosody $\mathbf{x}_p \in \mathbb{N}^{m \times L}$, content $\mathbf{x}_c \in \mathbb{N}^{n \times L}$, and acoustic tokens $\mathbf{x}_a \in \mathbb{N}^{k \times L}$, along with a speaker embedding $\mathbf{s} \in \mathbb{R}^{D_s}$, where $m$, $n$, $k$ are the number of Residual Vector Quantization (RVQ) codebooks (each with vocabulary size $v$), $L$ is the token sequence length, and $D_s$ is the speaker embedding dimension. The Codec Decoder reconstructs the speech waveform from these tokens and the speaker embedding.

\paragraph{Text Input}
Given a text transcript $\mathbf{T}$, we use a grapheme-to-phoneme converter \cite{g2pE2019} to obtain the corresponding phoneme sequence, which is mapped into a sequence of tokens $\mathbf{P}$ consisting of $N$ tokens. A \textit{Phoneme Encoder} then transforms $\mathbf{P}$ into a sequence of embeddings $\mathbf{p} \in \mathbb{R}^{N \times D}$, where $D$ denotes the hidden dimension.

\paragraph{Video Input}
Given a silent video $\mathbf{V}$, we follow the extraction pipeline in \cite{hpmdubbing} to obtain the face-cropped video $\mathbf{V}_{\text{face}}$ and lips-cropped video $\mathbf{V}_{\text{lip}}$ and their corresponding face feature embeddings $\mathbf{v}_{\text{face}} \in \mathbb{R}^{F \times D_f}$ and lip feature embeddings $\mathbf{v}_{\text{lip}} \in \mathbb{R}^{F \times D_l}$. Here, $F$ is the number of frames, and $D_f$ and $D_l$ denote the face and lip feature dimensions, respectively. 
Since these extracted features exist in different representation spaces, we apply a 1D convolution to map them into a shared space: $\hat{\mathbf{v}}_{\text{face}} \in \mathbb{R}^{F \times D}$ and $\hat{\mathbf{v}}_{\text{lip}} \in \mathbb{R}^{F \times D}$.

\vspace{-2pt}
\subsection{Zero-shot TTS Pre-training} \label{sec:pretrained-tts}
\vspace{-2pt}
This stage aims to generate high-quality speech with natural prosody and expressive characteristics. To achieve this, we follow the zero-shot TTS setting, which aims to synthesize speech that faithfully reproduces the speech attributes of unseen speakers using only a few seconds of their reference speech. Let $\mathbf{S}^r$ denote the reference speech. Our objective is to build a framework that synthesizes $\tilde{\mathbf{S}}$ by generating prosody $\tilde{\mathbf{x}}_p$, content $\tilde{\mathbf{x}}_c$, and acoustic $\tilde{\mathbf{x}}_a$ discrete token sequences conditioned on $\mathbf{S}^r$ and $\mathbf{T}$. The key components of this stage are described below.

\paragraph{Content Modeling} Figure \ref{fig:framework} illustrates the content attribute modeled by a simple deterministic architecture (Teacher box). The phoneme sequence $\mathbf{p}$ is processed to predict $n$ representations (one per RVQ codebook of content). These representations are then passed through a \textit{Projection} to obtain a latent representation of content $\tilde{\mathbf{h}}_c \in \mathbb{R}^{n \times L \times D}$, and through a \textit{Content Head} to produce a probability distribution over content tokens, from which $\tilde{\mathbf{x}}_c$ is predicted.

\paragraph{Discrete Flow-based Prosody-Acoustic} For prosody and acoustic attributes, we design a discrete flow matching framework to model their joint distribution.  As shown in Figure \ref{fig:framework}b, the Discrete Flow Denoiser outputs the joint prosody-acoustic distribution, represented as a probabilistic velocity field. Our denoiser is built on a Diffusion Transformer (DiT) \cite{dit}, which takes the current denoising target $\mathbf{x}_t =  [\mathbf{x}_{p,t}; \mathbf{x}_{a,t}] \in \mathbb{N}^{(m+k) \times L}$ as input, where $[;]$ denotes channel-wise concatenation, and $\mathbf{x}_{p,t} \in \mathbb{N}^{m \times L}$ and $\mathbf{x}_{a,t} \in \mathbb{N}^{k \times L}$ are the masked prosody and acoustic token sequences at step $t \in [0,1]$, respectively. Here, $\mathbf{x}_0$ denotes the fully masked source sequences and $\mathbf{x}_1$ denotes the ground-truth token sequences. The denoiser is conditioned on $\mathbf{c}$, which comprises prosody tokens $\mathbf{x}^r_p \in \mathbb{N}^{m \times L'}$, acoustic tokens $\mathbf{x}^r_a \in \mathbb{N}^{k \times L'}$ (where $L'$ is the token sequence length of $\mathbf{S}^r$), and a speaker embedding $\mathbf{s}$ extracted from $\mathbf{S}^r$ via the Codec Encoder, along with the latent representation of content $\tilde{\mathbf{h}}_c$. These conditioning signals provide contextual information to predict the features in $\mathbb{R}^{(m + k) \times L \times v}$ that represent the posterior distributions for prosody and acoustic tokens. This design enables the model to simultaneously generate tokens that capture expressive prosody $\tilde{\mathbf{x}}_p$ and realistic acoustic characteristics $\tilde{\mathbf{x}}_a$, remaining consistent with the speaker's vocal identity. The objective of the DFPA module is to learn a probabilistic denoiser $p(\mathbf{x}_1 | \mathbf{x}_t, \mathbf{c})$ that reconstructs masked tokens under varying masking ratios. 
Formally, the model is trained with the Cross-Entropy loss:

\vspace{-0.5cm}
\begin{equation}
\mathcal{L}_{\text{DFM}} = 
-\sum_{i \in \mathcal{T}}
\mathbb{E}_{t \sim \mathcal{U}[0,1], (\mathbf{x}_0, \mathbf{x}_1), \mathbf{x}_t}
\left[
\log p_{1|t}(\mathbf{x}_1^i \mid \mathbf{x}_t, \mathbf{c}; \theta)
\right],
\label{eq:dfm_loss}
\end{equation}
where $p_{1|t}(\cdot \mid \mathbf{x}_t, \mathbf{c}; \theta)$ denotes the posterior distribution of $\mathbf{x}_1$ 
given the partially corrupted sequence $\mathbf{x}_t$, conditioning inputs $\mathbf{c}$, and $\theta$ represents the parameters of the denoiser.
Here, $\mathcal{T} = \{1, \ldots, (m + k)L\}$ represents the index set of token positions, 
$\mathbf{x}_t \sim p_t(\mathbf{x} \mid \mathbf{x}_0, \mathbf{x}_1)$ is sampled from the mixture path, 
and $\mathbf{x}_0 \sim p$, $\mathbf{x}_1 \sim q$ denote the source and target distributions, respectively.

Once trained, this TTS model serves as a teacher in our dubbing framework. Its pre-learned representations of pronunciation, prosody, and acoustic characteristics are transferred to the video-driven dubbing stage, providing strong initialization for naturalness, pronunciation accuracy, and expressive speech generation. We encourage readers to refer to the Supplementary Material for details on constructing the conditioning inputs, the model architecture, and training configurations of the zero-shot TTS pre-training stage.

\vspace{-2pt}
\subsection{Video Dubbing Adaptation}
\vspace{-2pt}
We detail the components used to adapt the pre-trained TTS model for video dubbing.
\subsubsection{Face-to-Prosody Mapper}
The FaPro is designed to bridge facial expressions and prosodic information in speech. In video dubbing, it is crucial that facial expressions primarily determine the prosody of the dubbed speech. Using this strong correlation, we develop FaPro to generate prosodic latent representations of speech directly from facial expressions. We first process the facial features $\hat{\mathbf{v}}_{\text{face}}$ using an upsampling layer, implemented as a learnable interpolation-convolution module, to expand the video sequence length to match that of the speech. Next, ConvNeXt V2 \cite{Woo_2023_CVPR} blocks are applied to enhance the temporal modeling capability of the resulting representations. Finally, we design a \textit{Prosody Predictor} composed of $m$ Feed-Forward Transformer (FFT) layers (one per RVQ codebook of prosody) to iteratively extracts representations for each residual codebook of prosody tokens, where each layer models dependencies conditioned on the outputs of the previous layers. We denote the resulting representation as the \textit{global prosody prior}: $
    \tilde{\mathbf{z}}_p = \text{FaPro}(\hat{\mathbf{v}}_{\text{face}}) \in \mathbb{R}^{m \times L \times D}.$
\subsubsection{Synchronizer} \label{sec:synchronizer}
To effectively bridge the modality gap between text, video, and speech, we propose the \textit{Synchronizer}. It enhances video-speech alignment through the dual-alignment mechanism: video-text alignment, and speech-text alignment, which together enforce robust cross-modal synchronization. In the following, we detail the key elements of this module.

\paragraph{Video-Text Alignment}
Leveraging the strong correlation between lip movements and phonemes, we capture cross-modal interactions using transformer blocks equipped with cross-attention, where lip features $\hat{\mathbf{v}}_{\text{lip}}$ serve as queries and phoneme features $\mathbf{p}$ as keys and values. From the final transformer block, we obtain hidden states $\mathbf{h}_{\text{vt}} \in \mathbb{R}^{F \times D}$, and an attention map $\mathcal{A}_{\text{VT}} \in \mathbb{R}^{F \times N}$. To explicitly enforce alignment, we apply a contrastive loss to the attention map $\mathcal{A}_{\text{VT}}$. The durations of the phonemes are obtained from the Montreal Forced Aligner (MFA) \cite{mfa}, converted into frame spans via the video frame rate, and used to construct a binary alignment matrix $\mathcal{M}_{\text{VT}} \in \mathbb{R}^{F \times N}$. Each entry $m_{ij}$ indicates whether the $i$-th video frame corresponds to the $j$-th phoneme. The alignment loss is defined as:
\vspace{-0.2cm}
\begin{equation}
\begin{aligned}
\scriptstyle
\mathcal{L}_{\text{VT}} = -\sum_{j=1}^{N} \log \frac{\sum_{i=1}^{F} m_{ij} \cdot \exp(a_{ij}/\tau)}{\sum_{i=1}^{F} \exp(a_{ij}/\tau)} -\sum_{i=1}^{F} \log \frac{\sum_{j=1}^{N} m_{ij} \cdot \exp(a_{ij}/\tau)}{\sum_{j=1}^{N} \exp(a_{ij}/\tau)},
\end{aligned}
\label{eq:loss_align1}
\end{equation}
where $a_{ij}$ denotes the attention score between the $i$-th frame and the $j$-th phoneme, and $\tau$ is the temperature coefficient.

\paragraph{Duration-Based Regulation}
We retain textual information by directly utilizing the attention weights obtained from Equation \eqref{eq:loss_align1}. Through \textit{Monotonic Alignment Search} (MAS) \cite{glow_tts}, these weights are converted into phoneme-to-frame durations, allowing each phoneme embedding to be duplicated according to its assigned span, yielding 
$\hat{\mathbf{p}} \in \mathbb{R}^{F \times D}$. The resulting expanded sequence is then projected through a linear transformation and added to the cross-modal hidden states $\mathbf{h}_{\text{vt}}$. We further upsample the resulting representation using a learnable interpolation-convolution module, expanding the sequence length to match the length of the speech features and projecting it into the latent speech space:
\vspace{-0.2cm}
\begin{equation}
    \mathbf{h}_s = \mathrm{Upsample}(\mathbf{h}_{\text{vt}} + \mathrm{Proj}(\hat{\mathbf{p}})) \in \mathbb{R}^{L \times D}.
    \label{eq:h_s}
\end{equation}
\paragraph{Speech-Text Alignment}
Although the previous step aligns the video and text modalities, minor misalignments can still occur after upsampling. To refine synchronization, we introduce an additional contrastive alignment between the phoneme embeddings and the speech representations. Similar to the video-text alignment, transformer blocks with cross-attention are employed, where the speech features $\mathbf{h}_s$ from Equation \eqref{eq:h_s} serve as queries and the phoneme embeddings $\mathbf{p}$ act as keys and values. From the final transformer block, we obtain hidden states $\mathbf{h}_{\text{st}} \in \mathbb{R}^{L \times D}$ and an attention map $\mathcal{A}_{\text{ST}} \in \mathbb{R}^{L \times N}$. A binary alignment matrix $\mathcal{M}_{\text{ST}} \in \mathbb{R}^{L \times N}$ is constructed analogously to $\mathcal{M}_{\text{VT}}$ by converting the phoneme durations into speech-token spans according to the sampling rate. The phoneme-speech contrastive loss is defined as:
\vspace{-0.2cm}
\begin{equation}
\begin{aligned}
\scriptstyle
\mathcal{L}_{\text{ST}} = -\sum_{j=1}^{N}\log \frac{\sum_{i=1}^{L} m_{ij} \cdot \exp(a_{ij}/\tau)}{\sum_{i=1}^{L} \exp(a_{ij}/\tau)} - \sum_{i=1}^{L}\log \frac{\sum_{j=1}^{N} m_{ij} \cdot \exp(a_{ij}/\tau)}{\sum_{j=1}^{N} \exp(a_{ij}/\tau)},
\end{aligned}
\label{eq:loss_align2}
\end{equation}
where $a_{ij}$ denotes the attention score between the $i$-th speech token and the $j$-th phoneme.

\paragraph{Refinement}
Finally, the representation $\mathbf{h}_\text{st}$ is refined with ConvNeXt V2 \cite{Woo_2023_CVPR} blocks, which offer strong temporal modeling capabilities for speech-related tasks. The resulting aligned representation, denoted as $\mathbf{h}_{\text{sync}} \in \mathbb{R}^{L \times D}$, captures both the implicit alignment between video and speech, as well as the explicit alignments between video-text and speech-text pairs.

\subsubsection{Content-Consistent Temporal Adaptation} \label{sec:ccta}
CCTA facilitates the transfer of semantic content knowledge from the TTS domain and aligns it with the video to ensure temporal synchronization while preserving semantic consistency. To achieve this, we adopt the content modeling architecture and initialize its weights from the pre-trained TTS model described in Section~\ref{sec:pretrained-tts}. During adaptation, we freeze the \textit{Projection and Content Head} and replace the duration prediction mechanism used in the TTS model with the Synchronizer, as detailed in Section~\ref{sec:synchronizer}.

To ensure semantic and linguistic consistency, we impose a distillation constraint between the teacher (TTS) model and the student (video dubbing) model. The distillation loss enforces feature alignment by ensuring consistency between intermediate representations before they are passed to the \textit{Content Predictor}. Specifically, we first fuse the global prosody representation $\tilde{\mathbf{z}}_p$ with the residual content information $\mathbf{h}_{\text{sync}}$. The integration is performed through a fusion layer composed of channel-wise concatenation followed by linear transformations. This simple yet effective mechanism leverages the inherent alignment between prosodic and linguistic features. The resulting fused representation is then passed through transformer blocks denoted as $\mathbf{z}_s \in \mathbb{R}^{L \times D}$ for the student feature. Similarly, we extract the corresponding representation from the teacher model before the content predictor, denoted as $\mathbf{z}_t \in \mathbb{R}^{L' \times D}$, where $L'$ represents the temporal length of the teacher module. To align these feature spaces, we employ a cosine similarity-based distillation loss:
\vspace{-0.3cm}
\begin{equation}
    \mathcal{L}_{\text{distill}} = \frac{1}{B} \sum_{i=1}^{B} \Big[ 1 - \cos\big(\phi(\mathbf{z}_t^{(i)}), \phi(\mathbf{z}_c^{(i)})\big) \Big],
    \label{eq:distill_loss}
\end{equation}
where $\phi(\cdot)$ denotes the average pooling function, and $B$ is the batch size. Finally, a \textit{Content Predictor}, sharing the same architecture as the \textit{Prosody Predictor} in the FaPro, iteratively extracts $n$ representations for each residual codebook of content tokens, denoted as $\mathbf{h}_{\text{p2c}} \in \mathbb{R}^{n \times L \times D}$, and passes them through a \textit{Content Head}, implemented as a Multi-Layer Perceptron (MLP), to predict logits for each residual codebook, optimized using a Cross-Entropy (CE) loss:
\vspace{-3pt}
\begin{equation}
    \mathcal{L}_c = \text{CE}(\mathbf{x}_c, \text{MLP}(\mathbf{h}_{\text{p2c}})),
    \label{eq:content_loss}
\end{equation}
where $\mathbf{x}_c$ denotes the target content sequence. Simultaneously, a projection layer produces a latent representation 
$\tilde{\mathbf{h}}_{\text{p2c}} = \text{Proj}(\mathbf{h}_{\text{p2c}}) \in \mathbb{R}^{n \times L \times D}$, 
which captures the prosody-to-content correlation. This representation provides contextual guidance for modeling the subsequent speech attributes.

\subsubsection{Discrete Flow-based Prosody-Acoustic Adaptation} \label{sec:DFPA_adapt}
To effectively inherit the capability of the DFPA module in capturing expressive prosody and realistic acoustic characteristics learned from large-scale TTS pre-training, we initialize the denoiser with the pre-trained weights and fine-tune it with several conditioning context modifications. Particularly, we replace $\mathbf{x}^r_p$ with $\tilde{\mathbf{z}}_{p}$, set $\mathbf{x}^r_a$ to \textit{zero embeddings}, and substitute $\tilde{\mathbf{h}}_c$ with $\tilde{\mathbf{h}}_{\text{p2c}}$. These signals are incorporated as control inputs, enabling DFPA to generate diverse yet globally coherent prosody tokens alongside acoustic tokens. We further elaborate on the construction of the conditioning context $\mathbf{c}$ and describe how it is integrated into our framework.

\paragraph{Contextual Modeling} 
Given the current denoising target token sequences $\mathbf{x}_t$, which are obtained by concatenating the prosody token sequences $\mathbf{x}_{p,t}$ and the acoustic token sequences $\mathbf{x}_{a,t}$, we first apply dedicated embedders to encode them into latent representations: $
\mathbf{e}_{p,t} = \mathbf{E}_p(\mathbf{x}_{p,t}) \in \mathbb{R}^{m \times L \times D}, \mathbf{e}_{a,t} = \mathbf{E}_a(\mathbf{x}_{a,t}) \in \mathbb{R}^{k \times L \times D}$. We then integrate the $\tilde{\mathbf{h}}_{\text{p2c}}$ obtained from CCTA stage by concatenation to form the complete set of attributes, including prosody, content, and acoustic representations: $
\mathbf{e}_t = [\mathbf{e}_{p,t}; \tilde{\mathbf{h}}_{\text{p2c}}; \mathbf{e}_{a,t}] \in \mathbb{R}^{(m+n+k) \times L \times D}$. We then construct a conditioning representation $\mathbf{e}_{\text{cond}} = [\tilde{\mathbf{z}}_{p}; \mathbf{0}; \mathbf{0}] \in \mathbb{R}^{(m+n+k) \times L \times D}$, where $\tilde{\mathbf{z}}_{p}$ serves as the conditioning signal to promote diverse and expressive prosody generation, and \textit{zero embeddings} occupy the content and acoustic slots. Finally, this process yields the complete input $
\mathbf{e} = [\mathbf{e}_{\text{cond}};\mathbf{e}_t] \in \mathbb{R}^{(m+n+k) \times 2L \times D}
$. The input $\mathbf{e}$, together with a global conditioning vector $\mathbf{c}_g \in \mathbb{R}^D$ formed by the summation of the speaker embedding $\mathbf{s}$ projected onto $\mathbb{R}^D$ and timestep embedding $\mathbf{t} \in \mathbb{R}^D$, is fed into the denoiser, which predicts features in $\mathbb{R}^{(m + k) \times L \times v}$ that represent the posterior distributions of the prosody and acoustic tokens, as described in Section \ref{sec:pretrained-tts}.

\vspace{-2pt}
\subsection{Training Objectives}
\vspace{-2pt}
The total loss of dubbing $\mathcal{L}_{\text{Dubbing}}$ integrates all objectives introduced in 
Equations~\eqref{eq:dfm_loss},
\eqref{eq:loss_align1}, \eqref{eq:loss_align2}, \eqref{eq:distill_loss}, and
\eqref{eq:content_loss}.
In addition, following previous work \cite{emodubber, aligndit, choi25c_interspeech}, we introduce an auxiliary Connectionist Temporal Classification (CTC) loss to encourage the output of the \textit{Synchronizer}, $\mathbf{h}_{\text{sync}}$, to retain richer linguistic information. The overall loss is defined as:
\vspace{-0.1cm}
\begin{equation}
    \mathcal{L}_{\text{Dubbing}} =  
    \lambda_1 \mathcal{L}_{c} + 
    \lambda_2 \mathcal{L}_{\text{CTC}} +
    \lambda_3 \mathcal{L}_{\text{distill}} + 
    \lambda_4 \mathcal{L}_{\text{DFM}} + \mathcal{L}_{\text{align}},
    \label{eq:total_loss}
\end{equation}
where $\mathcal{L}_{\text{align}} = \lambda_5\mathcal{L}_{\text{VT}} +\lambda_6\mathcal{L}_{\text{ST}}$ and 
$\lambda_1$--$\lambda_6$ are weighting coefficients that balance the contribution of each term.

\begin{table*}
\centering
\caption{Performance on the \textit{Chem} dataset under both Setting 1.0 and Setting 2.0. In the Dub 1.0 Setting, the ground truth audio is used as the reference, while in the Dub 2.0 Setting, a non-ground-truth audio sample from the same speaker within the dataset is used as the reference. The best and second-best results are shown in \textbf{bold} and \underline{underlined}, respectively.}
\label{tab:overall-results-chem}
\resizebox{\textwidth}{!}{
\begin{tabular}{l|ccccc|cccccc}
\toprule[1.25pt]
\multirow{3}{*}{\textbf{Model}} & \multicolumn{5}{c|}{\textit{Setting 1.0}} & \multicolumn{6}{c}{\textit{Setting 2.0}} \\
\cmidrule(lr){2-6} \cmidrule(lr){7-12}
& \textbf{LSE-C} $\uparrow$ & \textbf{LSE-D} $\downarrow$ & \textbf{WER} $\downarrow$ & \textbf{SECS} $\uparrow$ & \textbf{UTMOS} $\uparrow$ & \textbf{LSE-C} $\uparrow$ & \textbf{LSE-D} $\downarrow$ & \textbf{WER} $\downarrow$ & \textbf{SECS} $\uparrow$ & \textbf{MOS-S} $\uparrow$ & \textbf{UTMOS} $\uparrow$ \\
\midrule
Ground Truth & 8.12 & 6.59 & 3.85 & 100.00 & 4.18 & 8.12 & 6.59 & 3.85 & 100.00 & 4.34 $\pm$ 0.18 & 4.18 \\
\midrule
V2C-Net \citep{v2c-net} \textcolor{gray}{CVPR'22} & 1.97 & 12.17 & 90.47 & 51.52 & 1.81 & 1.82 & 12.09 & 94.59 & 44.19 & - & 1.76 \\
HPMDubbing \citep{hpmdubbing} \textcolor{gray}{CVPR'23} & 7.85 & 7.19 & 16.05 & 85.09 & 2.16 & 3.98 & 9.50 & 29.82 & 73.55 & - & 2.01 \\
StyleDubber \citep{styledubber} \textcolor{gray}{ACL'24}& 3.87 & 10.92 & 13.14 & \underline{87.72} & 3.14 & 3.74 & 11.00 & 14.18 & \underline{82.07} & 4.02 $\pm$ 0.24 & 3.04 \\
Speaker2Dubber \citep{speaker2dubber} \textcolor{gray}{ACM MM'24}& 3.76 & 10.56 & 16.98 & 74.73 & 3.61 & 3.45 & 11.17 & 18.10 & 69.28 & 3.69 $\pm$ 0.25 & 3.64 \\
ProDubber \citep{produbber} \textcolor{gray}{CVPR'25} & 2.58 & 12.54 & \textbf{9.45} & 72.13 & \underline{3.85} & 2.78 & 12.14 & \underline{11.69} & 50.85 & 3.48 $\pm$ 0.26 & \underline{3.76} \\
EmoDubber \citep{emodubber} \textcolor{gray}{CVPR'25}& \underline{8.11} & \underline{6.92} & 11.72 & \textbf{90.62} & 3.82 & \underline{8.09} & \underline{6.96} & 12.81 & \textbf{85.06} & 4.14 $\pm$ 0.23 & 3.75 \\
\midrule
\rowcolor{aliceblue}
\textbf{DiFlowDubber (Ours)} & \textbf{8.31} & \textbf{6.73} & \underline{9.65} & 84.59 & \textbf{4.02} & \textbf{8.37} & \textbf{6.70} & \textbf{11.12} & 73.93 & \textbf{4.18 $\pm$ 0.18} & \textbf{4.03} \\
\bottomrule[1.25pt]
\end{tabular}
}
\end{table*}
\begin{table*}
\centering
\caption{Performance on the \textit{GRID} dataset is evaluated under the same two Dub settings as in the \textit{Chem} dataset.}
\label{tab:overall-results-grid}
\resizebox{\textwidth}{!}{
\begin{tabular}{l|ccccc|ccccccc}
\toprule[1.25pt]
\multirow{3}{*}{\textbf{Model}} & \multicolumn{5}{c|}{\textit{Setting 1.0}} & \multicolumn{6}{c}{\textit{Setting 2.0}} \\
\cmidrule(lr){2-6} \cmidrule(lr){7-12}
& \textbf{LSE-C} $\uparrow$ & \textbf{LSE-D} $\downarrow$ & \textbf{WER} $\downarrow$ & \textbf{SECS} $\uparrow$ & \textbf{UTMOS} $\uparrow$ & \textbf{LSE-C} $\uparrow$ & \textbf{LSE-D} $\downarrow$ & \textbf{WER} $\downarrow$ & \textbf{SECS} $\uparrow$ & \textbf{MOS-S} $\uparrow$  & \textbf{UTMOS} $\uparrow$ \\
\midrule
Ground Truth & 7.13 & 6.78 & 22.41 & 100.00 & 3.94 & 7.13 & 6.78 & 22.41 & 100.00 & 4.58 $\pm$ 0.13 & 3.94 \\
\midrule
V2C-Net \citep{v2c-net} \textcolor{gray}{CVPR'22} & 5.59 & 9.52 & 47.82 & 80.98 & 2.41 & 5.34 & 9.76 & 49.09 & 71.51 & - & 2.40 \\
HPMDubbing \citep{hpmdubbing} \textcolor{gray}{CVPR'23} & 5.76 & 9.13 & 45.51 & 85.11 & 2.14 & 5.82 & 9.10 & 44.15 & 71.99 & - & 2.11 \\
StyleDubber \citep{styledubber} \textcolor{gray}{ACL'24}& 6.12 & 9.03 & 18.88 & \underline{93.79} & 3.73 & 6.09 & 9.08 & 19.58 & \textbf{86.67} & 3.83 $\pm$ 0.24 & 3.71 \\
Speaker2Dubber \citep{speaker2dubber} \textcolor{gray}{ACM MM'24}& 5.27 & 9.84 & \underline{17.07} & \textbf{94.50} & 3.69 & 5.19 & 9.93 & \underline{17.42} & 85.76 & 3.57 $\pm$ 0.26 & 3.73 \\
ProDubber \citep{produbber} \textcolor{gray}{CVPR'25}& 5.23 & 9.59 & 18.60 & 89.03 & \underline{3.87} & 5.56 & 9.37 & 19.17 & 81.72 & 4.04 $\pm$ 0.16 & \underline{3.86} \\
EmoDubber \citep{emodubber} \textcolor{gray}{CVPR'25}& \underline{7.12} & \underline{6.82} & 18.53 & 92.22 & 3.83 & \underline{7.10} & \underline{6.89} & 19.75 & \underline{86.02} & 4.33 $\pm$ 0.17 & 3.81 \\
\midrule
\rowcolor{aliceblue}
\textbf{DiFlowDubber (Ours)} & \textbf{7.32} & \textbf{6.73} & \textbf{16.79} & 82.52 & \textbf{3.95} & \textbf{7.28} & \textbf{6.78} & \textbf{16.21} & 71.63 & \textbf{4.38 $\pm$ 0.18} & \textbf{3.93} \\
\bottomrule[1.25pt]
\end{tabular}
}
\end{table*}
\section{Experiments} \label{sec:experiments}
\vspace{-2pt}
\subsection{Experimental Setup}
\vspace{-2pt}
\paragraph{Datasets} 
For zero-shot TTS pre-training stage, we use the \textbf{LibriTTS} dataset \cite{libritts}.
In the video dubbing adaptation stage, our model is trained and evaluated on the \textbf{Chem}~\cite{neural_dubber} and \textbf{GRID} \cite{GRID} datasets. Additional details
are provided in the \textit{Dataset Details} section of the Supplementary Material.

\paragraph{Evaluation Metrics} We employ comprehensive objective metrics to evaluate the synthesized speech. To assess temporal alignment between synthesized speech and visual content, we adopt Lip-Sync Error Confidence (LSE-C) and Lip-Sync Error Distance (LSE-D), computed using embeddings extracted from a pretrained SyncNet model \citep{syncnet}. For speech quality evaluation, we measure pronunciation accuracy using Word Error Rate (WER), computed with Whisper-V3 \citep{whisper} as the automatic speech recognition system. Additionally, we use Speaker Encoder Cosine Similarity (SECS), following \citep{styledubber, speaker2dubber}, to quantify speaker identity consistency between the generated speech and the reference audio. Naturalness is evaluated using UTMOS \citep{utmos}. To assess efficiency, we report the Real-Time Factor (RTF) under matched conditions, using a single NVIDIA A100 80GB GPU and excluding video preprocessing, thereby demonstrating the computational efficiency of our discrete flow matching framework. 

\paragraph{Implementation Details \& Baselines}
Please refer to the \textit{Implementation Details} section of the Supplementary Material for a comprehensive description. Additionally, we compare our model with previous dubbing baselines, with
further information provided in the \textit{Baselines Details} section of the Supplementary Material.

\vspace{-2pt}
\subsection{Main Results}
\vspace{-6pt}
\paragraph{Result on the Chem dataset} As shown in Table \ref{tab:overall-results-chem}, 
our proposed method with 128 function evaluations (NFEs) outperforms current SOTA models 
across most metrics on the Chem benchmark. DiFlowDubber achieves superior results in LSE-C 
and LSE-D, indicating stronger synchronization and demonstrating the effectiveness of our 
Synchronizer in enhancing temporal alignment between linguistic content and lip movements. 
In terms of pronunciation accuracy, our model achieves the best performance in Setting 2.0 
and ranks second in Setting 1.0. Although ProDubber attains competitive WER scores in both 
settings, it performs poorly in lip synchronization, as reflected by its weak LSE scores, 
suggesting behavior closer to a TTS system rather than a true dubbing model. For speaker 
similarity, DiFlowDubber offers no clear advantage over the baselines in SECS, which we 
attribute to the use of the same speaker embedding extractor for both baseline speech 
generation and SECS computation, introducing bias in favor of those systems. To further 
verify the true perceptual similarity, we conduct a subjective evaluation following the 
Mean Opinion Score (MOS) protocol under Setting 2.0. Specifically, we use MOS-Similarity 
(MOS-S) to evaluate speaker similarity. In this evaluation, 20 listeners rate 30 randomly 
selected synthesized speech samples on a 1-5 scale based on perceived similarity to the 
reference speech. The results show that our model achieves the highest MOS-S score among 
the four latest baselines, highlighting that our method achieves superior speaker 
faithfulness where it matters most perceptually. Regarding naturalness and speech quality, as measured by UTMOS, our method also achieves SOTA results, further validating its 
ability to generate natural and expressive dubbed speech.

\paragraph{Result on the GRID dataset} Table \ref{tab:overall-results-grid} reports the performance of our method with 128 NFEs compared to baseline systems. Our approach consistently achieves SOTA performance across most metrics, including synchronization (LSE-C, LSE-D), pronunciation clarity (WER), and speech quality (UTMOS), in both dubbing settings, with the exception of the SECS metric. For SECS, we assume this trend follows the same pattern as previously discussed in the Chem benchmark; nevertheless, our method achieves the highest MOS-S score. These results further confirm the effectiveness of our design choices in improving synchronization, pronunciation accuracy, and speech quality.
\vspace{-2pt}
\subsection{Qualitative Analysis}
\vspace{-2pt}
\begin{table}
\centering
\caption{Ablation study results on the Chem test set.}
\label{tab:ablation-results}
\resizebox{\columnwidth}{!}{
\begin{tabular}{l|lcccccc}
\toprule[1.25pt]
\textbf{\#} & \textbf{Model} & \textbf{AVSync} $\uparrow$ & \textbf{LSE-C} $\uparrow$ & \textbf{LSE-D} $\downarrow$ & \textbf{WER} $\downarrow$ & \textbf{SECS} $\uparrow$ & \textbf{UTMOS} $\uparrow$ \\
\midrule
\rowcolor{aliceblue}
1 & \textbf{Ours} & \textbf{0.721} & 8.31 & 6.73 & \textbf{9.65} & \textbf{84.59} & \textbf{4.02}\\
\midrule
\multicolumn{5}{l}{\textit{Zero-shot TTS Pre-training}} \\
2 & \emph{~~w/o} Zero-shot TTS Pre-training & 0.715 & 8.17 & 6.79 & 12.04 & 83.72 & 3.53\\
\midrule
\multicolumn{5}{l}{\textit{Synchronizer}} \\
3 & \emph{~~w/o} ($\mathcal{L}_{\text{VT}} + \mathcal{L}_{\text{ST}}$) & 0.652 & 8.26 & 6.77 & 17.15 & 84.29 & 3.90\\
4 & \emph{~~w/o} $\mathcal{L}_{\text{VT}}$ & 0.681 & 8.36 & 6.71 & 16.57 & 83.94 & 3.87\\
5 & \emph{~~w/o} $\mathcal{L}_{\text{ST}}$ & 0.689 & 8.31 & 6.71 & 12.60 & 84.46 & 3.93\\
\midrule
\multicolumn{5}{l}{\textit{Content-Consistent Temporal Adaptation}} \\
6 & \emph{~~w/o} $\mathcal{L}_{\text{distill}}$ & 0.713 & 8.33 & 6.70 & 12.62 & 84.23 & 3.97 \\
7 & \emph{~~w/o} $\mathcal{L}_{\text{CTC}}$ & 0.703 & 8.35 & 6.67 & 12.28 & 84.35 & 3.99 \\
8 & \emph{~~w/o} Fuse Global Prosody Prior & 0.694 & 8.38 & 6.69 & 10.43 & 79.10 & 3.99 \\
\midrule
\multicolumn{5}{l}{\textit{Facial Features}} \\
9 & \emph{~~w/o} FaPro & 0.691 & 8.37 & 6.65 & 11.89 & 83.83 & 3.95\\
\bottomrule[1.25pt]
\end{tabular}
}
\end{table}

\begin{table}
\centering
\caption{Analysis of NFE in inference under the Dub 1.0 Setting. \textbf{\#Params} exclude video-related encoders.}
\label{tab:effect_nfe}
\resizebox{\columnwidth}{!}{
\begin{tabular}{c|cc|ccccccc}
\toprule[1.25pt]
\textbf{Model} & \textbf{\#Params} & \textbf{NFE} & 
\textbf{RTF} $\downarrow$ & 
\textbf{LSE-C} $\uparrow$ & 
\textbf{LSE-D} $\downarrow$ & 
\textbf{WER} $\downarrow$ &
\textbf{SECS} $\uparrow$ & 
\textbf{UTMOS} $\uparrow$ \\
\midrule

ProDubber & 200M & 5  & 0.08 & 2.58 & 12.54 & 9.45 & 72.13 & 3.85 \\
EmoDubber & 116M & 10 & 0.05 & 8.11 & 6.92 & 11.72 & 90.62 & 3.82 \\
\midrule

\multirow{8}{*}{\textbf{DiFlowDubber}} & 
\multirow{8}{*}{250M} 

& 1   & 0.03 & 8.25 & 6.78 & 10.72 & 83.63 & 2.82 \\
& & 2   & 0.04 & 8.27 & 6.74 & 10.99 & 83.34 & 2.79 \\
& & 4   & 0.04 & 8.32 & 6.73 & 10.79 & 83.82 & 3.50 \\
& & \cellcolor{aliceblue}8   & \cellcolor{aliceblue}0.05 & \cellcolor{aliceblue}8.29 & \cellcolor{aliceblue}6.73 & \cellcolor{aliceblue}10.28 & \cellcolor{aliceblue}84.50 & \cellcolor{aliceblue}3.85 \\
& & \cellcolor{aliceblue}10   & \cellcolor{aliceblue}0.06 & \cellcolor{aliceblue}8.30 & \cellcolor{aliceblue}6.72 & \cellcolor{aliceblue}10.82 & \cellcolor{aliceblue}83.73 & \cellcolor{aliceblue}3.88 \\
& & 16  & 0.08 & 8.27 & 6.75 & 11.02 & 83.83 & 3.94 \\
& & 32  & 0.12 & 8.34 & 6.69 & 10.90 & 84.58 & 4.00 \\
& & 64  & 0.22 & 8.32 & 6.70 & 10.79 & 84.07 & 4.03 \\
& & 128 & 0.40 & 8.31 & 6.73 & 9.65  & 84.59 & 4.02 \\

\bottomrule[1.25pt]
\end{tabular}
}
\vspace{-0.5cm}
\end{table}
\begin{figure}[t]
    \vspace{-0.2cm}
    \centering
    \includegraphics[width=0.88\columnwidth]{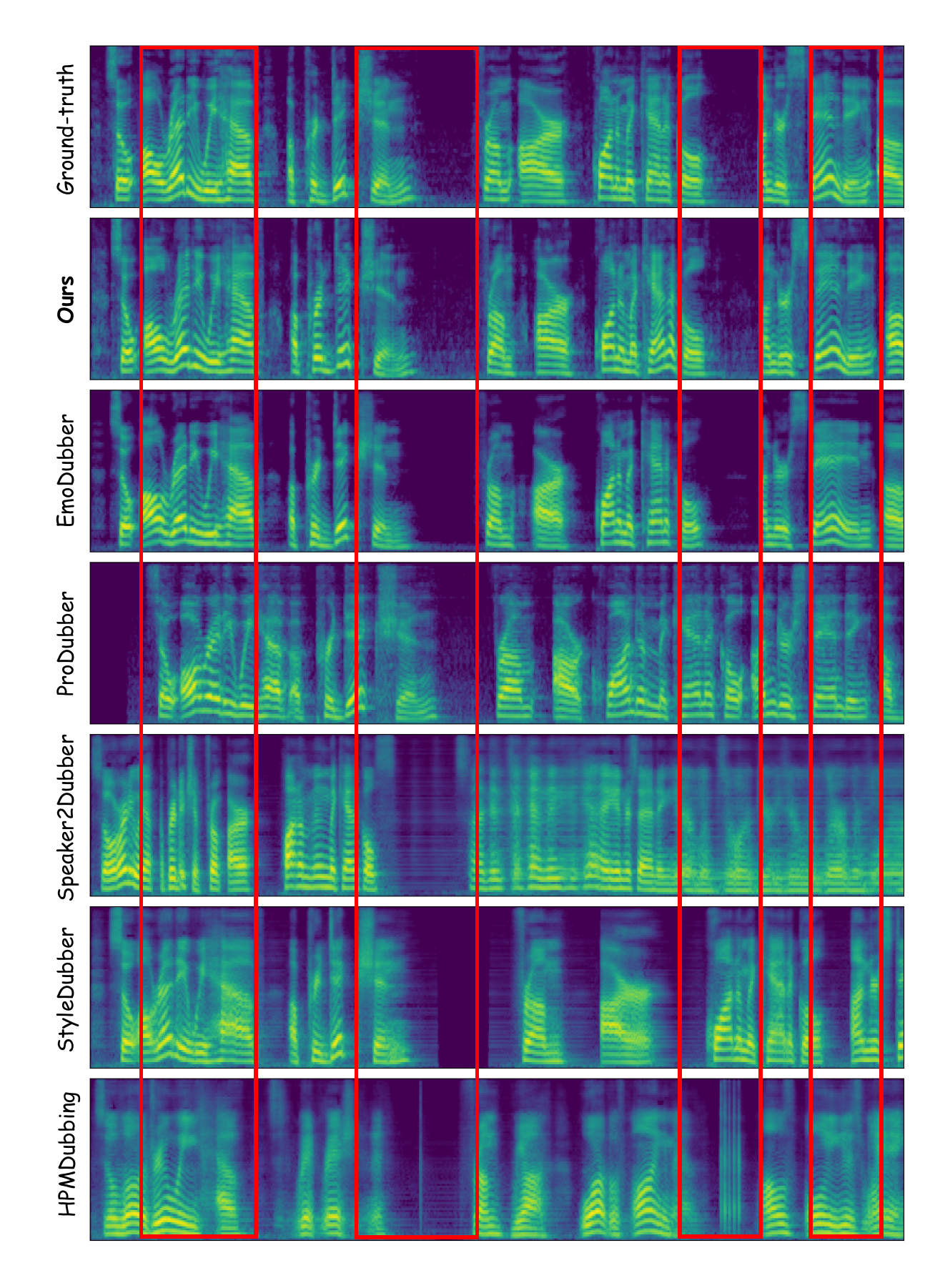}
    \caption{Mel-spectrogram visualization compared with Ground-Truth speech.}
    \label{fig:visualization}
    \vspace{-0.5cm}
\end{figure}
Figure~\ref{fig:visualization} provides a qualitative comparison of the mel-spectrograms generated in different dubbing systems. The highlighted regions (in red boxes) indicate areas where different models show noticeable discrepancies in speech quality. Across all samples, DiFlowDubber produces spectrograms most similar to the ground truth, preserving clear harmonic structure, accurate temporal boundaries, and natural prosodic dynamics. In addition, it effectively maintains pitch continuity and alignment with lip movements in the corresponding video frames, confirming that our model produces speech with more natural rhythm, precise synchronization, and expressive delivery.
\vspace{-0.7cm}
\subsection{Ablation Study}
\vspace{-2pt}
We perform ablation studies under the Dub 1.0 setting of the Chem benchmark using 128 NFEs. To more comprehensively evaluate the contribution of the Synchronizer module, we additionally report AVSync \citep{aligndit} along with LSE in the standard benchmark.

\paragraph{Effect of Zero-shot TTS Pre-training} Table \ref{tab:ablation-results} shows that without the zero-shot TTS pre-training, the generated dubbing speech exhibits a significant reduction in naturalness and pronunciation accuracy (see row \#2), demonstrating that our approach effectively inherits the expressive prosody and acoustic quality learned from large-scale TTS corpus while maintaining precise lip synchronization during dubbing.

\paragraph{Effect of Synchronizer} Table \ref{tab:ablation-results} shows that when the model is not constrained by cross-modal alignment (see row \#3), it performs the worst in both synchronization and pronunciation accuracy, highlighting the importance of alignment constraints for performance improvement. From rows \#4 and \#5, we observe that using $\mathcal{L}_{\text{ST}}$ loss is the primary driver for synchronization, significantly raising the AVSync score, while $\mathcal{L}_{\text{VT}}$ improves both synchronization and pronunciation accuracy due to the inherent correlation between lip motion and text. Thus, combining both losses yields a more balanced and overall superior performance (see row \#1).

\paragraph{Effect of Content-Consistent Temporal Adaptation} As shown in Table \ref{tab:ablation-results}, we observe that removing the $\mathcal{L}_{\text{distill}}$ loss (see row \#6) leads to degraded pronunciation accuracy, demonstrating the importance of effectively transferring knowledge from the TTS domain, where pronunciation is highly accurate, to the video dubbing task while maintaining temporal lip-sync alignment. When the $\mathcal{L}_{\text{CTC}}$ loss is removed (see row \#7), a similar pattern is observed, highlighting its role in facilitating more accurate alignment learning between the input text and the output of the Synchronizer. From row \#8, fusing global prosody prior with content information enhances speaker similarity, indicating that face-derived global prosody encodes speaker-dependent cues essential for perceptually consistent dubbing.

\paragraph{Effect of FaPro} Table~\ref{tab:ablation-results} shows that removing the FaPro module (see row \#9) results in consistent quality degradation across almost all metrics,  confirming the importance of global prosody cues derived from facial expressions. These results indicate that FaPro effectively captures expressive facial dynamics to guide prosody generation.

\paragraph{Effect of NFE} We conducted experiments with varying numbers of NFEs during inference on the Chem test set, as shown in Table \ref{tab:effect_nfe}. We further compare the RTF of our method with prior diffusion- and flow-based dubbing systems (ProDubber,  EmoDubber). The results indicate that increasing the NFE substantially improves the UTMOS score, while other metrics remain largely stable. Performance remains consistent between 32 and 128 NFEs; notably, DiFlowDubber achieves competitive results across metrics and maintains an RTF equivalent to EmoDubber (0.05) at only 8 (RTF = 0.05) or 10 (RTF = 0.06) NFEs, despite a larger parameter count. These findings further demonstrate the advantage of the discrete flow matching framework over its counterparts based on continuous flow matching and diffusion in achieving high-quality speech with fewer sampling steps.
\vspace{-5pt}
\section{Conclusion}
In this paper, we present \textbf{DiFlowDubber}, the first end-to-end video dubbing framework built upon a discrete flow matching backbone. Our two-stage pipeline first pre-trains a zero-shot TTS system, where a simple deterministic architecture captures linguistic structure and the DFPA module models expressive prosody and realistic acoustic characteristics from large-scale TTS corpora. In the video dubbing adaptation stage, the CCTA module transfers content knowledge from the TTS domain via the Synchronizer, which enforces precise cross-modal synchronization through a dual-alignment mechanism over text, video, and speech. Meanwhile, the FaPro extracts global prosodic cues from facial expressions. These cues, together with the outputs of the Synchronizer, are fused to capture prosody-content correlations, thereby guiding the DFPA to generate expressive prosody and acoustic tokens for content-consistent speech. Extensive experimental results validate the effectiveness of our approach.

{\small
\bibliographystyle{ieeenat_fullname}
\bibliography{main}
}

\maketitlesupplementary
\appendix

\section*{Summary}
This appendix contains additional materials for the paper \textit{``DiFlowDubber: Discrete Flow Matching for Automated Video Dubbing via Cross-Modal Alignment and Synchronization"}. The appendix is organized as follows:
\begin{itemize}
    \item Section \ref{sec:dataset} presents dataset details, including LibriTTS for zero-shot TTS pre-training and the Chem and GRID benchmarks for video dubbing.
    \item Section \ref{sec:implement} provides implementation details, covering preprocessing, model architectures for both the zero-shot TTS pre-training stage and the video dubbing adaptation stage, as well as training configurations.
    \item Section \ref{sec:baselines} describes the baseline methods used for comparison.
    \item Section \ref{sec:additional_results} reports additional qualitative results, including alignment visualizations of the \textit{Synchronizer} and mel-spectrogram comparisons with baseline systems.
    \item Section \ref{sec:limitations} discusses the limitations of the proposed method.
\end{itemize}

\section{Dataset Details} \label{sec:dataset}
We adopt a two-stage training process. In the zero-shot TTS pre-training stage, we train on the LibriTTS dataset, while in the video dubbing adaptation stage, we use the Chem and GRID benchmarks.

\paragraph{LibriTTS}
The LibriTTS \cite{libritts} is a large-scale multi-speaker English speech corpus designed for text-to-speech (TTS) research. It contains recordings from numerous speakers with diverse accents, vocal styles, and speaking rates. Each utterance is carefully segmented and aligned with its corresponding text transcript. In our work, we use 470 hours of the LibriTTS dataset as the text-speech corpus for the zero-shot TTS pre-training stage.

\paragraph{Chem}
The Chem dataset \cite{neural_dubber} consists of classroom lecture videos featuring a chemistry instructor, collected from publicly available YouTube recordings~\citep{neural_dubber}. It spans approximately nine hours of content and is segmented at the sentence level to facilitate precise dubbing alignment. The dataset contains 6,082 samples for training, 50 for validation, and 196 for testing.

\paragraph{GRID}
The GRID corpus \cite{GRID} is a standard benchmark for evaluating multi-speaker video dubbing systems. It features recordings from 33 distinct speakers, each contributing 1,000 short English utterances. All recordings were captured under controlled studio conditions with a consistent visual background. The dataset includes 32,670 training instances and 3,280 test instances.

\section{Implementation Details} \label{sec:implement}
\subsection{Preprocessing}
We use FACodec~\cite{ns3} as the audio tokenizer, detokenizer, and speaker extractor. Audio is sampled at 16 kHz and tokenized using FACodec at 80 tokens/s, while videos are sampled at 25 FPS, resulting in a fixed 5:16 length ratio between video frames and speech tokens. Each lip-cropped frame is resized to $96 \times 96$ pixels. The numbers of quantizers are $m = 1$ for prosody, $n = 2$ for content, and $k = 3$ for acoustic tokens, each with a vocabulary size of 1024. To obtain the ground-truth alignment matrices for video-text and speech-text alignment, we first extract the duration of each phoneme using the Montreal Forced Aligner (MFA) \cite{mfa}. Based on these durations, we construct the video-text alignment matrix by multiplying each phoneme's duration by the video frame rate (25 FPS) to determine the number of frames corresponding to each phoneme. Similarly, the speech-text alignment matrix is constructed by multiplying the phoneme durations by the token rate (80 tokens/s) to obtain the number of speech tokens aligned with each phoneme.

\subsection{Model Details}
\subsubsection{Zero-shot TTS Pre-training Stage}
\paragraph{Content Modeling} Following conventional duration-based alignment modules \cite{fastspeech, fastspeech2, kim2023pflow}, our architecture consists of a \textit{Phoneme Encoder}, a \textit{Duration Predictor}, a \textit{Length Regulator}, and \textit{Feed-Forward Transformer} (FFT) layers. We model the content by adapting existing duration-alignment mechanisms to the discrete token modeling setting. Specifically, we employ the duration predictor from \cite{fastspeech} to estimate the duration (i.e., the number of tokens) for each input phoneme. Each phoneme is then duplicated by the length regulator based on the predicted durations and passed through 2 FFT blocks to refine the representations. Each FFT block contains 4 attention heads, a hidden size of 256, an output dimension of 768, convolutional filter sizes of 1024 with kernel sizes $[9, 1]$, a dropout rate of 0.2, and a maximum sequence length of 5000. After the FFT blocks, the \textit{Content Predictor} employs two FFT blocks with the same configuration to generate two textual representations that capture progressively richer features through the stacked FFT layers. The final output consists of two branches: the first is projected through a linear layer to obtain the \textit{final hidden state}, while the second is projected onto the vocabulary space to produce \textit{logits} representing the distribution of content tokens.

\paragraph{Discrete Flow Matching Overview} The objective of Discrete Flow Matching is to transform source samples $\mathbf{x}_0 \sim p$ into target samples $\mathbf{x}_1 \sim q$. In our formulation, the source distribution corresponds to all-mask tokens \texttt{[MASK]}, while the target distribution $\mathbf{x}_1 = [\mathbf{x}_{p,1};\mathbf{x}_{a,1}] \in \mathbb{N}^{(m+k) \times L}$ is factorized into prosodic $\mathbf{x}_{p,1} \in \mathbb{N}^{m \times L}$ and acoustic $\mathbf{x}_{a,1} \in \mathbb{N}^{k \times L}$ components, enabling structured joint learning. Following \cite{dfm}, we employ a monotonic scheduler $\kappa_t \in [0,1]$ with boundary conditions $\kappa_0 = 0$ and $\kappa_1 = 1$, where $t \in [0,1]$ denotes continuous time. In our implementation, we set $\kappa_t = t^2$. This scheduler progressively interpolates between the source and target distributions, smoothly transitioning from $\mathbf{x}_0$ to $\mathbf{x}_1$ as $\kappa_t$ increases. We then construct a conditional probability path, referred to as the \textit{mixture path}, which linearly interpolates between the source and target distributions: $
p_t(\mathbf{x}^i | \mathbf{x}_0, \mathbf{x}_1) = (1 - \kappa_t)\,p_0(\mathbf{x}^i | \mathbf{x}_0) + \kappa_t\,p_1(\mathbf{x}^i | \mathbf{x}_1).
$
This path defines an evolution of discrete states governed by a probability velocity field $\mathbf{u}_t$, expressed as:
\begin{equation}
\label{eq:u_t}
\mathbf{u}^i_t(\mathbf{x}^i,\mathbf{x}_t) = \frac{\dot{\kappa}_t}{1-\kappa_t}\Big[p_{1|t}(\mathbf{x}^i|\mathbf{x}_t, \mathbf{c}; \theta) - p_t(\mathbf{x}^i | \mathbf{x}_t)\Big],
\end{equation}
where $\dot{\kappa}_t$ denotes the time derivative of the scheduler, $\theta$ represents the learnable parameters of the probability denoiser, and $p_{1|t}(\cdot|\mathbf{x}_t, \mathbf{c}; \theta)$ is the posterior distribution of $\mathbf{x}_1$ given the partially corrupted sequence $\mathbf{x}_t$ and the conditioning context $\mathbf{c}$.

\paragraph{Denoiser Architecture}
We use a Diffusion Transformer (DiT) \cite{dit} as the denoiser in our discrete flow matching framework. The DiT consists of 8 layers with a hidden size of 768 and 8 attention heads, and is enhanced with rotary position embeddings (RoPE) \cite{su2024roformer}. The time step is embedded using an MLP that maps it to a 768-dimensional vector, while the speaker embedding, originally of dimension 256, is linearly projected to the same 768-dimensional space. The global conditioning vector $\mathbf{c}_g$ is then formed by summing the resulting the time embedding $\mathbf{t}$ and the speaker embedding $\mathbf{s}$ projected via an MLP, which are used to condition the Adaptive Layer Normalization (AdaLN) layers. We employ a long skip connection from the input to the output of the final Transformer block. The final prediction stage performs a non-affine layer normalization followed by AdaLN modulation and a linear projection. The global conditioning vector is processed through a SiLU-activated \cite{elfwing2018sigmoid} MLP to generate shift and scale parameters that modulate the normalized hidden states. The final linear projection produces an output of $(1+3) \times 768$ dimensions, corresponding to one prosody quantizer and three acoustic quantizers.

\paragraph{Contextual Modeling} We describe the construction of the input to the \textit{Denoiser}. The input consists of two parts: the conditioning input, denoted as $\mathbf{e}_{\text{cond}}$, and the current denoising target, $\mathbf{e}_t$ at time step $t$.

For $\mathbf{e}_{\text{cond}}$, given a reference speech sample $\mathbf{S}^r$, we use FACodec \cite{ns3} to extract discrete speech tokens along with a speaker embedding, yielding
$\mathbf{x}^r_p \in \mathbb{N}^{m \times L'}$,
$\mathbf{x}^r_c \in \mathbb{N}^{n \times L'}$,
$\mathbf{x}^r_a \in \mathbb{N}^{k \times L'}$, and
$\mathbf{s} \in \mathbb{R}^{D_s}$,
representing the prosody, content, and acoustic token sequences, and the speaker embedding, respectively.
Here, $m$, $n$, and $k$ denote the number of Residual Vector Quantization (RVQ) codebooks for prosody, content, and acoustic representations, each with a vocabulary size of $v$. $L'$ is the token sequence length, and $D_s$ is the speaker embedding dimension. We pass $\mathbf{x}^r_p$ and $\mathbf{x}^r_a$ to dedicated embedders to encode them into latent representations
$\mathbf{e}^r_p \in \mathbb{R}^{m \times L' \times D}$ and
$\mathbf{e}^r_a \in \mathbb{R}^{k \times L' \times D}$.
In the zero-shot TTS setting, which aims to mimic the speaker style of the reference speech, we omit the content tokens $\mathbf{x}^r_c$ and replace them with \textit{zero embeddings}. The final conditioning vector is defined as
$\mathbf{e}_{\text{cond}} = [\mathbf{e}^r_p;\mathbf{0}; \mathbf{e}^r_a]  \in \mathbb{R}^{(m+n+k) \times L' \times D}$.

For $\mathbf{e}_t$, given the current denoising target tokens $\mathbf{x}_t = [\mathbf{x}_{p,t};\mathbf{x}_{a,t}] \in \mathbb{N}^{(m+k) \times L}$, we feed $\mathbf{x}_{p,t} \in \mathbb{N}^{m \times L}$ and $\mathbf{x}_{a,t} \in \mathbb{N}^{k \times L}$ into the dedicated embedders to obtain latent representations: $\mathbf{e}_{p,t} = \mathbf{E}_p(\mathbf{x}_{p,t}) \in \mathbb{R}^{m \times L \times D}$ and
$\mathbf{e}_{a,t} =  \mathbf{E}_a(\mathbf{x}_{a,t}) \in \mathbb{R}^{k \times L \times D}$.
Since speech rhythm also depends on linguistic content \cite{zuo2025rhythm}, we incorporate the content information from the \textit{final hidden state} of the content modeling module, denoted as $\tilde{\mathbf{h}}_c$.
The final denoising target is then defined as
$\mathbf{e}_t = [\mathbf{e}_{p,t};\tilde{\mathbf{h}}_c;\mathbf{e}_{a,t}] \in \mathbb{R}^{(m + n + k) \times L \times D}$.

Finally, the complete input to the \textit{Denoiser} is constructed by concatenating $\mathbf{e}_{\text{cond}}$ and $\mathbf{e}_t$ along the temporal dimension:
$\mathbf{e} = [\mathbf{e}_{\text{cond}}; \mathbf{e}_t] \in \mathbb{R}^{(m + n + k) \times (L' + L) \times D}$. The input $\mathbf{e}$ is permuted to shape $\mathbb{R}^{(L' + L) \times (m + n + k)D}$, projected to $\mathbb{R}^{(L' + L) \times D}$, and fed into the \textit{Denoiser}. The resulting output is then passed through a final transformation layer comprising layer normalization, AdaLN-based modulation conditioned on $\mathbf{c}_g$, and a linear projection, yielding features of shape $\mathbb{R}^{(L' + L) \times (m+k)D}$. We discard the conditioning portion and permute the result to obtain the final hidden representation in $\mathbb{R}^{(m + k) \times L \times D}$, which is then passed through a dedicated MLP to produce the posterior distributions over prosody and acoustic tokens in $\mathbb{R}^{(m+k) \times L \times v}$.

\paragraph{Total Loss} The overall training objective in this stage consists of three components. First, we optimize the \textit{Duration Predictor} using the Mean Squared Error loss on the logarithmic scale, denoted as $\mathcal{L}_d$. Second, $\mathcal{L}_c$ corresponds to the content modeling loss, computed as the Cross-Entropy between the predicted logits and the target content sequence. Third, we employ the discrete flow matching loss, $\mathcal{L}_{\text{DFM}}$ (described in the main paper). The overall loss is formulated as follow:
\begin{equation}
\mathcal{L}_{\text{TTS}} = \lambda_1 \mathcal{L}_d + \lambda_2 \mathcal{L}_c + \lambda_3 \mathcal{L}_{\text{DFM}},
\label{eq:total_loss_tts}
\end{equation}
where $\lambda_1 = 0.5$, $\lambda_2 = 1.0$, and $\lambda_3 = 1.0$ are the weighting coefficients that balance the contribution of each term.

\subsubsection{Video Dubbing Adaptation Stage}
\paragraph{FaPro Module}
The architecture is composed of a learnable upsampling layer, a ConvNeXt V2 \cite{Woo_2023_CVPR} encoder stack, layer normalization, and a Transformer decoder. The upsampling layer applies linear interpolation followed by 4 sets of learnable convolutional transformations. Each set consists of a 1D convolution (kernel size 3, stride 1, padding 1), a group normalization layer with 1 group, and a Mish activation \cite{misra2019mish} function. A final 1D convolution projects the features to the target hidden dimension of 256. The upsampled features are then processed by a stack of 8 ConvNeXt V2 \cite{Woo_2023_CVPR} blocks, each with hidden dimension 256 and intermediate dimension 1024. Every block contains a depthwise 1D convolution with kernel size 7, followed by layer normalization and 2 linear projections. Between these projections, we apply a GELU activation \cite{hendrycks2016gelu} and a Global Response Normalization (GRN) module, which stabilizes the feature scale by normalizing the response of each channel with respect to its global L2 magnitude. A residual connection is added around each block, enabling stable and efficient learning of temporal dynamics across long facial sequences. After the ConvNeXt V2 \cite{Woo_2023_CVPR} encoder stack, a layer normalization layer is applied before forwarding the features to one Transformer decoder that serves as the \textit{Prosody Predictor}, built on the FFT block architecture. The decoder uses 4 attention heads, a hidden dimension of 256, and a feed-forward expansion ratio of 4, and operates with a maximum sequence length of 5000.

\paragraph{Synchronizer Module}
\begin{figure}[t]
    \centering
    \includegraphics[width=\columnwidth]{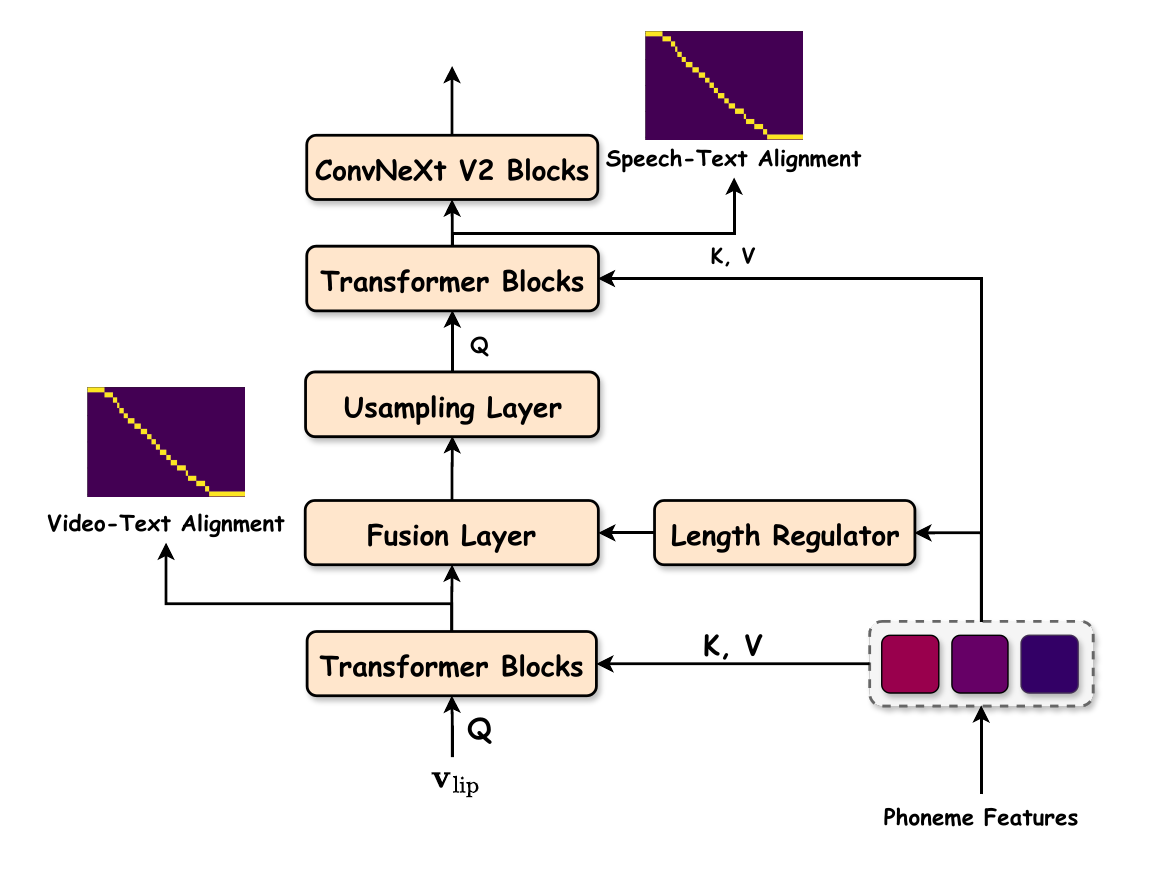}
    \caption{The detailed architecture of \textit{Synchronizer}.}
    \label{fig:synchronizer}
\end{figure}
As shown in Figure \ref{fig:synchronizer}, the module consists of 2 Transformer stacks, a learnable upsampling layer, a light fusion network, a ConvNeXt V2 \cite{Woo_2023_CVPR} encoder stack, and two layer normalization layers. We use 8 Transformer blocks for both video-text and speech-text alignment, where the attention mechanism follows a Monotonic Multi-Head Attention \cite{Ma2020Monotonic} formulation with 4 attention heads, a hidden dimension of 256, and an intermediate dimension of 1024. The learnable upsampling layer and the ConvNeXt V2 \cite{Woo_2023_CVPR} encoder stack use the same architecture and hyperparameters as those in the \textit{FaPro} module. The light fusion network is implemented as a linear projection layer.

\subsection{Training Details}
The training is conducted in 2 stages:

\paragraph{Zero-shot TTS Pre-training}
We train on 470 hours of the LibriTTS dataset \cite{libritts} using 4 NVIDIA A100 80GB GPUs with a batch size of 16 for a total of 300k training steps. The learning rate is set to $10^{-4}$ with a 10\% warm-up ratio and a weight decay of 0.01, optimized using the AdamW optimizer \cite{loshchilov2018decoupled}. The weighting coefficients $\lambda_1$, $\lambda_2$, and $\lambda_3$ are defined as in Equation \ref{eq:total_loss_tts}.

\paragraph{Video Dubbing Adaptation}
Training is performed on the Chem \cite{neural_dubber} and GRID \cite{GRID} datasets using 4 NVIDIA A100 80GB GPUs with batch sizes of 16 and 64, for 150 and 200 epochs, respectively. We use the AdamW optimizer \cite{loshchilov2018decoupled} with a learning rate of $10^{-3}$, a weight decay of 0.01, and a 5\% warm-up ratio. The weights for the loss $\mathcal{L}_{\text{Dubbing}}$ (defined in the main paper) are set to $\lambda_1 = 1.0$, $\lambda_2 = 0.1$, $\lambda_3 = 0.1$, $\lambda_4 = 1.0$, $\lambda_5 = 0.001$, and $\lambda_6 = 0.001$.

\section{Baselines Details} \label{sec:baselines}
We compare our model with previous dubbing systems, including:

\begin{itemize}
    \item \textbf{V2C-Net}~\cite{v2c-net} is the baseline model for the V2C task, which generates speech from text while conditioning on both reference audio and video. It fuses visual emotion cues with text and voice identity to produce speech that matches the target speaker and reflects video-driven affect.
    \item \textbf{HPMDubbing}~\cite{hpmdubbing} is a hierarchical prosody modeling framework that bridges video features (lip, face, scene) and speech prosody to improve emotional alignment and timing in video dubbing.
    \item \textbf{StyleDubber}~\cite{styledubber} switches dubbing learning from the frame level to the phoneme level, using a multimodal style adaptor and phoneme-guided lip aligner to jointly model pronunciation style and visual emotion while maintaining lip-sync.
    \item \textbf{Speaker2Dubber}~\cite{speaker2dubber} is a two-stage dubbing method that first pre-trains a phoneme encoder on large-scale TTS data to learn clear pronunciation, then adapts it to video dubbing with prosody and duration consistency learning.
    \item \textbf{ProDubber}~\cite{produbber} introduces prosody-enhanced acoustic pre-training and acoustic-disentangled prosody adapting to fully leverage TTS pre-training, improving acoustic quality and prosody control in dubbing.
    \item \textbf{EmoDubber}~\cite{emodubber} is an emotion-controllable dubbing architecture that uses lip-related prosody aligning, pronunciation enhancing, and a flow-based emotion controller to achieve high-quality lip sync, intelligibility, and user-specified emotion type and intensity.
\end{itemize}
\begin{figure*}[t]
    \centering
    \includegraphics[width=\textwidth]{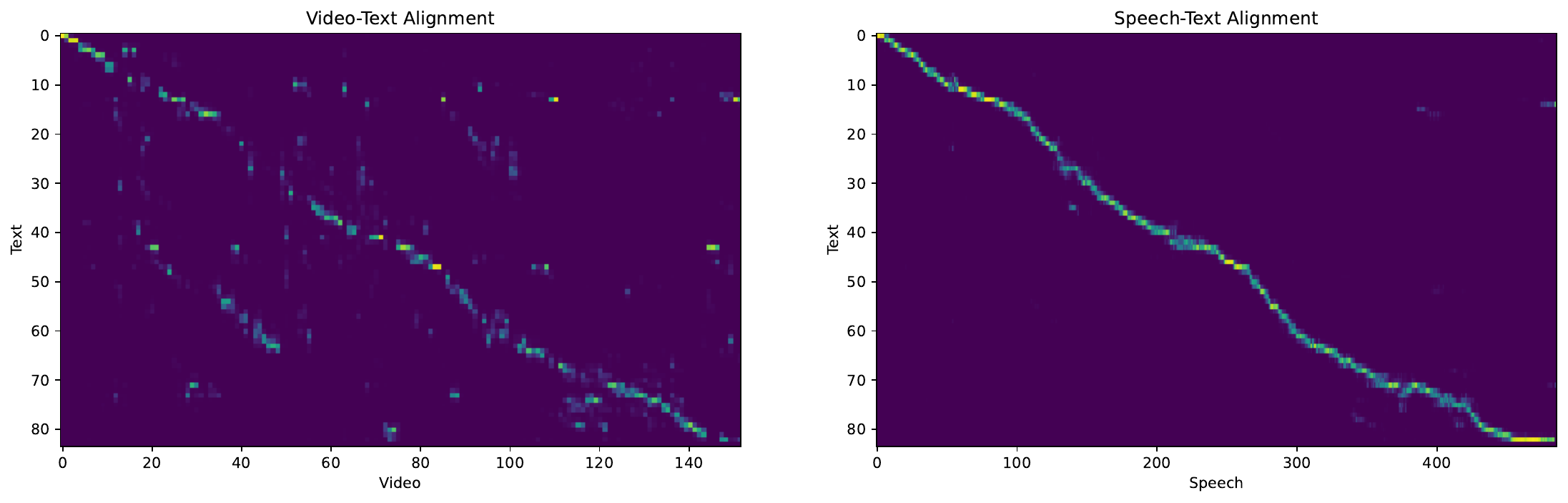}
    \caption{Alignment visualization of the \textit{Synchronizer}. Left: Video-Text alignment matrix. Right: Speech-Text alignment matrix.}
    \label{fig:alignment}
\end{figure*}
\begin{figure*}[t]
    \centering
    \includegraphics[width=\textwidth]{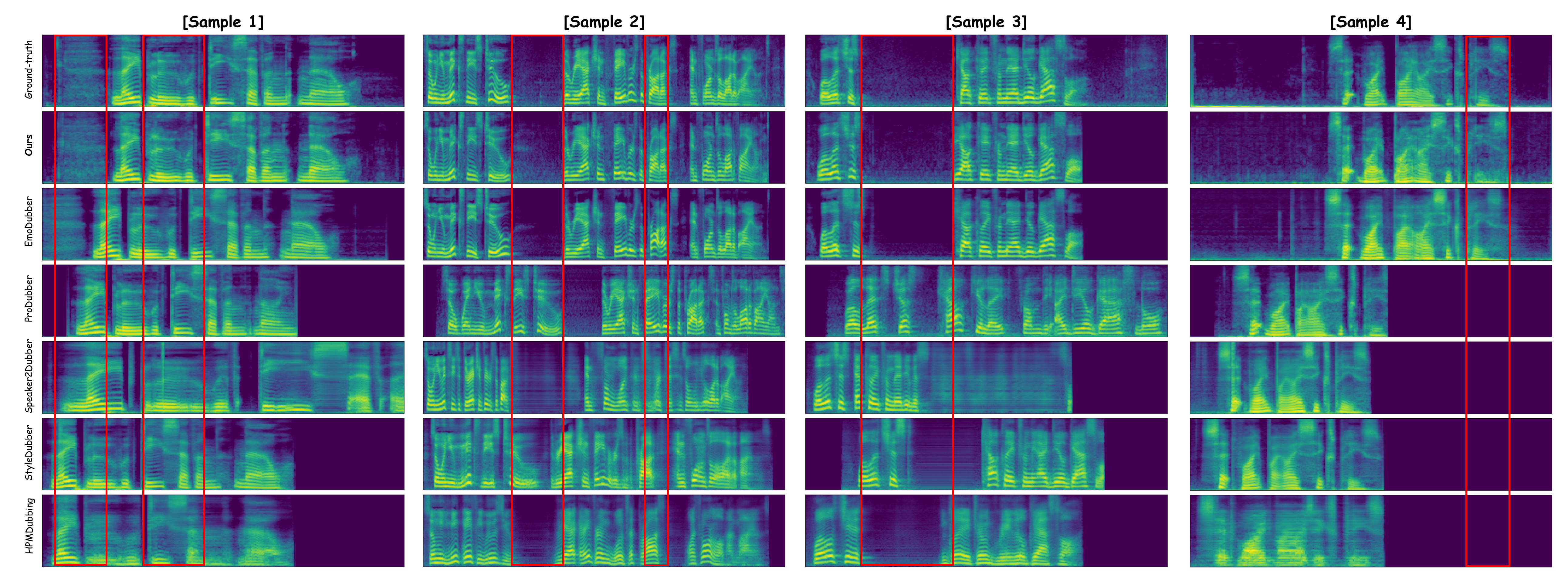}
    \caption{Qualitative comparison of mel-spectrograms. Our method produces spectra closest to the ground-truth, with clearer harmonics, cleaner pauses, and better temporal synchronization than baseline methods.}
    \label{fig:additional_vis}
\end{figure*}

\section{Additional Qualitative Results} \label{sec:additional_results}
\subsection{Alignment Visualization of Synchronizer}
Figure~\ref{fig:alignment} visualizes the attention maps learned by the \textit{Synchronizer} module. The left panel shows the video-text alignment between lip-frame features and phoneme embeddings, while the right panel shows the speech-text alignment between discrete speech tokens and phonemes. In both cases, the attention concentrates along a clear diagonal, indicating monotonic and fine-grained temporal alignment. This confirms that the Synchronizer successfully learns to align visual and speech streams with the textual content, providing a reliable cross-modal bridge that underpins the strong lip synchronization and pronunciation accuracy observed in our experiments.

\subsection{Mel-spectrogram Visualization}
We provide additional qualitative comparisons in Figure~\ref{fig:additional_vis}. For 4 representative samples, we visualize the mel-spectrograms of the ground-truth speech, our DiFlowDubber, and all baseline methods. The red bounding boxes highlight regions where different models exhibit noticeable differences in speech quality. Across all samples, our method produces spectra most similar to the ground-truth, preserving clear harmonic structure and temporal dynamics. The voiced regions in our results align well with the visual boundaries, whereas baseline systems often exhibit temporal drift and over-smoothed harmonics, resulting in degraded synchronization.
\subsection{Expressive Metrics for Prosody Validation}
To effectively validate that the DFPA module generates diverse yet globally consistent prosody under the guidance of the FaPro module, we conduct additional evaluations using a set of expressive prosody metrics that directly measure pitch modeling accuracy and emotional consistency:
\begin{itemize}
    \item \textbf{Gross Pitch Error (GPE)} \citep{10.1109/ICASSP.2009.4960497}: Measures the percentage of voiced frames where the relative error between the fundamental frequency F0 of synthesized speech and the ground truth exceeds a predefined threshold (typically 20\%), indicating major pitch deviations.
    \item \textbf{Voicing Decision Error (VDE)} \citep{10.1109/ICASSP.2009.4960497}: Computes the percentage of frames with incorrect voiced/unvoiced decisions relative to the reference, reflecting rhythmic consistency.
    \item \textbf{F0 Frame Error (FFE)} \citep{10.1109/ICASSP.2009.4960497}: Combines GPE and VDE, representing the percentage of frames with either gross pitch error or incorrect voicing decisions, thereby summarizing overall $F_0$ modeling accuracy.
    \item \textbf{Emotion Similarity (Emo-SIM)}: Measures emotional consistency by computing the cosine similarity between emotion embeddings extracted from the synthesized and reference speech using a pretrained emotion recognition model. In our experiments, we adopt the Emo2Vec\footnote{\url{https://github.com/ddlBoJack/emotion2vec}} model for embedding extraction.
\end{itemize}
\begin{table}
\centering
\caption{Evaluation of prosodic expressiveness and emotional consistency on the Chem dataset (Setting 2.0).}
\label{tab:expressive}
\resizebox{0.9\columnwidth}{!}{
\begin{tabular}{c|cccc}
\toprule[1.25pt]
\textbf{Model} & \textbf{FFE}$\downarrow$ & \textbf{GPE}$\downarrow$ & \textbf{VDE}$\downarrow$ & \textbf{Emo-SIM}$\uparrow$ 
\\
\midrule
HPMDubbing & 0.535 & 0.473 & 0.289 & 0.979 \\
StyleDubber & 0.583 & 0.493 & 0.360 & 0.976 \\
Speaker2Dubber & 0.639 & 0.519 & 0.408 & 0.979 \\
ProDubber & 0.653 & 0.562 & 0.436 & 0.959 \\
EmoDubber & 0.426 & 0.408 & 0.220 & 0.977 \\
\midrule
\rowcolor{aliceblue}
\textbf{DiFlowDubber (ours)} & \textbf{0.395} & \textbf{0.361} & \textbf{0.209} & \textbf{0.983} \\
\bottomrule[1.25pt]
\end{tabular}
}
\end{table}
Since the global prosody prior is derived from facial expressions, the target expressive intent is implicitly encoded in the corresponding ground-truth speech. Therefore, we use the ground-truth audio directly as the reference when computing these metrics, enabling us to measure how faithfully the synthesized speech aligns with the intended prosodic and emotional characteristics. As shown in Table \ref{tab:expressive}, we conduct the evaluation under Setting 2.0 to simulate a more challenging scenario for expressive dubbing. The results show that DiFlowDubber achieves the best performance across all expressive metrics. Specifically, our model significantly reduces pitch and voicing errors compared to strong baselines like EmoDubber (0.395 vs. 0.426 FFE), while achieving the highest emotional consistency (0.983 Emo-SIM). These findings confirm that leveraging facial dynamics as a prosody prior effectively guides the DFPA module to generate speech that is both prosodically and emotionally consistent with the visual content.
\section{Limitations}
\label{sec:limitations}
Although our proposed method significantly enhances the quality of generated dubbing, it still faces certain limitations. The framework depends on the third-party FACodec \cite{ns3}, from which it inherits certain constraints. In future work, we plan to adapt our system to operate with alternative codec models. Furthermore, the current design does not fully meet expectations in voice cloning. Developing more effective approaches to mimic speaker timbre from audio in real-world video dubbing remains an open area for improvement.

\end{document}